# Machine Learning in Artificial Intelligence: Towards a Common Understanding

Niklas Kühl
Karlsruhe Institute of Technology
kuehl@kit.edu

Marc Goutier
Karlsruhe Institute of Technology
marc.goutier@kit.edu

Robin Hirt
Karlsruhe Institute of Technology
hirt@kit.edu

Gerhard Satzger
Karlsruhe Institute of Technology
gerhard.satzger@kit.edu

## Abstract

*The application of "machine learning" and "artificial intelligence" has become popular within the last decade. Both terms are frequently used in science and media, sometimes interchangeably, sometimes with different meanings. In this work, we aim to clarify the relationship between these terms and, in particular, to specify the contribution of machine learning to artificial intelligence. We review relevant literature and present a conceptual framework which clarifies the role of machine learning to build (artificial) intelligent agents. Hence, we seek to provide more terminological clarity and a starting point for (interdisciplinary) discussions and future research.*

## 1. Introduction

In his US senate hearing in April 2018, Mark Zuckerberg stressed the necessary capabilities of Facebook's "AI tools (…) to (…) identify hate speech (…)" or " (…) terrorist propaganda" [1]. Researchers would typically describe such tasks of identifying specific instances within social media platforms as *classification tasks* within the field of *(supervised) machine learning* [2]–[4]. However, with rising popularity of *artificial intelligence (AI)* [5], the term AI is often used interchangeably with machine learning–not only by Facebook's CEO in the example above or in other interviews [6], but also across various theoretical and application-oriented contributions in recent literature [7]–[9]. Carner (2017) even states that he still uses AI as a synonym for machine learning although knowing this is not correct [10]. Such ambiguity, though, may lead to multiple imprecisions both in research and practice when conversing about methods, concepts, and results.

It seems surprising that despite of the frequent use of the terms, there is hardly any helpful scientific delineation. Thus, this paper aims to shed light on the relation of the two terms *machine learning* and *artificial intelligence*. We elaborate on the role of machine learning within instantiations of artificial intelligence, precisely within intelligent agents. To do so, we take a machine learning perspective on the capabilities of intelligent agents as well as the corresponding implementation.

The contribution of our paper is threefold. First, we expand the theoretical framework of Russel & Norvig (2015) [11] by further detailing the "thinking" layer of any intelligent agent by splitting it into separate "learning" and "executing" sublayers. Second, we show how this differentiation enables us to distinguish different contributions of machine learning for intelligent agents. Third, we draw on the implementations of the execution and learning sublayers ("backend") to define a continuum between human involvement and agent autonomy.

In the remainder of this paper, we first review relevant literature in the fields of machine learning and artificial intelligence. Next, we present and elaborate our conceptual framework which highlights the contribution of machine learning to artificial intelligence. On that basis, we derive an agenda for future research and conclude with a summary, current limitations, as well as an outlook.

## 2. Related work

As a base for our conceptual work, we first review the different notions, concepts, or definitions of machine learning and artificial intelligence within extant research. In addition, we elaborate in greater detail on the theories which we draw upon in our framework.

### 2.1. Terminology

Machine learning and artificial intelligence, as well as the terms data mining, deep learning and statistical learning are related, often present in the same context and sometimes used interchangeably. While the terms are common in different communities, their particular usage and meaning varies widely.

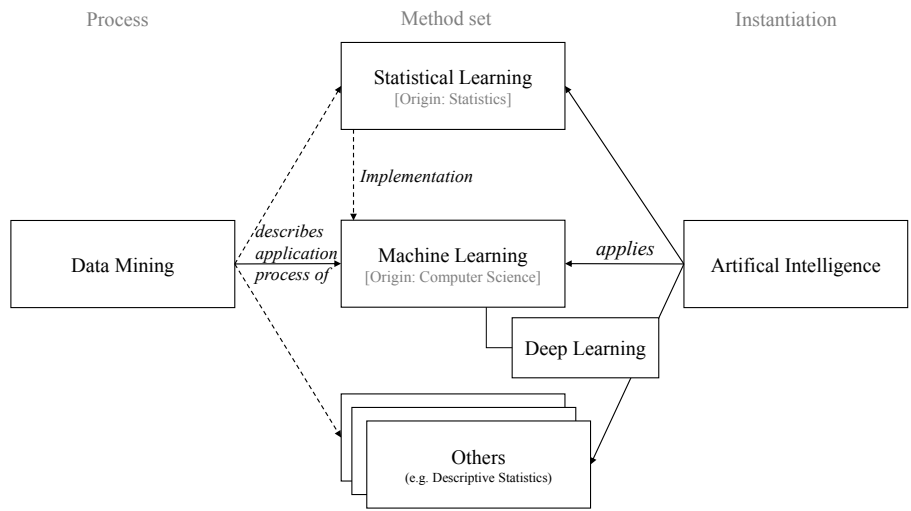

**Figure 1. General terminology used in this paper**

For instance, in the field of statistics the focus is on *statistical learning*, which is defined as a set of methods and algorithms to gain knowledge, predict outcomes, and make decisions by constructing models from a data set [12]. From a statistics point of view, machine learning can be regarded as an implementation of statistical learning [13].

Within the field of computer science, *machine learning* has the focus of designing efficient algorithms to solve problems with computational resources [14]. While machine learning utilizes approaches from statistics, it also includes methods which are not entirely based on previous work of statisticians—resulting in new and well-cited contributions to the field [15], [16]. Especially the method of deep learning raised increased interest within the past years [17]. *Deep learning* models are composed of multiple processing layers which are capable of learning representations of data with multiple levels of abstraction. Deep learning has drastically improved the capabilities of machine learning, e.g. in speech [18] or image recognition [19].

In demarcation to the previous terms, *data mining* describes the process on how to apply quantitative analytical methods, which help to solve real-world problems, e.g. in business settings [20]. In the case of machine learning, data mining is the process of generating meaningful machine learning models. The goal is not to develop further knowledge about machine learning algorithms, but to apply them to data in order to gain insights. Machine learning can therefore be seen as a foundation for data mining [21].

In contrast, *artificial intelligence* applies techniques like machine learning, statistical learning or other techniques like descriptive statistics to mimic intelligence in machines.

Figure 1 and the terms defined within this paragraph lay the foundation of the remainder of this work. However, the overall terminology and relationships of the concepts is discussed controversially [22]. Therefore, the focus of this paper is to bring more insight to the terminology and more precisely, to clarify the role of machine learning within AI. To gain a broader understanding for the terms machine learning and AI, we examine both in further detail.

## 2.2. Machine learning

Machine learning describes a set of techniques that are commonly used to solve a variety of real-world problems with the help of computer systems which can learn to solve a problem instead of being explicitly programmed [23]. In general, we can differentiate between unsupervised and supervised machine learning. For the course of this work, we focus on the latter, as the most-widely used methods are of supervised nature [24]. With regard to supervised machine learning, *learning* means that a series of examples ("past experience") is used to build knowledge about a given task [25]. Although statistical methods are used during the learning process, a manual adjustment or programming of rules or strategies to solve a problem is not required. In more detail, (supervised) machine learning techniques always aim to build a model by applying an algorithm on a set of known data points to gain insight on an unknown set of data [11], [26].

Thus, the processes of "creation" of a machine learning model slightly vary in their definition of phases but typically employ the three main phases of model initiation, performance estimation and deployment [27]: During the model initiation phase, a human user defines a problem, prepares and processes a data set and chooses a suitable machine learning algorithm for the given task. Then, during the performance estimation, various parameter permutations describing the algorithm are validated and a well-performing configuration is selected with respect to its performance in solving a specific task. Lastly, the model is deployed and put into practice to solve the task on unseen data.

Learning in general depicts a key facet of a human's cognition which "refers to all processes by which the sensory input is transformed, reduced, elaborated, stored, recovered, and used" [28, p. 4]. Humans process a vast amount of information by utilizing abstract knowledge that helps us to better understand incoming input. Due to their adaptive nature, machine learning models are able to mimic the cognitive abilities of a human being in an isolated manner.

However, machine learning solely represents a set of methods that enable to learn patterns in existing data, thus generating analytical models that can be utilized inside larger IT artifacts.

### 2.3. Artificial intelligence

The topic of artificial intelligence (AI) is rooted in different research disciplines, such as computer science [18, 19], philosophy [20, 21], or futures studies [22, 23]. In this work, we mainly focus on the field of computer science, as it is the most relevant one in identifying the contribution of machine learning to AI and in differentiating both terms.

AI research can be separated into different research streams [11]. These streams differ on the one hand as to the objective of AI application (*thinking* vs. *acting*), on the other hand as to the kind of decision making (targeting a *human-like decision* vs. an *ideal, rational decision*). This distinction leads to four research currents which are depicted in Table 1.

According to the "Cognitive Modeling" (i.e. thinking humanly) stream, an AI must be a machine with a mind [34]. This also includes performing human thinking [35], not only based on the same output as a human when given the same input, but also on the same reasoning steps which led to the very conclusion [36].

The "Laws of Thought" stream (i.e. thinking rationally) requires an AI to arrive at the rational decision despite what a human might answer.

**Table 1. AI research streams based on Russell & Norvig [11]**

| Objective / Application to | Humanly | Rationally |
|---|---|---|
| Thinking | Cognitive Modeling | "Laws of thought" |
| Acting | Turing Test | Rational Agent |

Therefore, an AI must follow the laws of thought by using computational models [37] which reflect logic.

The "Turing Test" (i.e. acting humanly) stream implies that an AI must act intelligently when interacting with humans. To accomplish these tasks, an AI must perform human tasks at least as good as humans [38]. These requirements can be tested by the Turing Test [39].

Finally, the "Rational Agent" stream considers an AI as a rational [11] or intelligent [40] agent[1]. This agent does not only act autonomously but also with the objective to achieve the rationally ideal outcome.

An alternative way to delineate AI is defining intelligence in general and using the resulting insights to create intelligent machines. Legg and Hutter [41] use intelligence tests, theories of human intelligence and psychological definitions to define a measurement of intelligence. Based on their definition, they use an agent-environment framework to describe intelligence in general and—in case the agent is a machine—artificial intelligence in particular. Their framework exhibits many similarities to the "acting rationally" stream.

Besides defining AI in general, the classification of AI is another topic in the field of AI research. Searle [42] suggests differentiating between weak and strong AI. Whereas a *weak AI* only pretends to think, a *strong AI* is a mind with mental states. Gubrud [43] however categorizes AI by taking the type of task into account. An *artificial general intelligence* (AGI) is an AI which in general, i.e. in any domain, acts at least on the same level as a human brain, however without requiring

---

[1] In this case, the terms *rational* and *intelligent* are used interchangeably in related work [11],[23]

consciousness. In contrast, a *narrow AI* is an AI that rivals or exceeds the human brain only in specific, limited tasks [44].

In the following, we will look into the "Rational Agent" stream in some more detail as it is of importance when regarding implementation of machine learning within AI. We will come back to the other three research streams in section 3 where we show that they are compatible with our framework of an agent-based AI.

According to the "Rational Agent" stream, the intelligence itself is manifested by the acting of agents. These agents are characterized by five features, namely they "operate autonomously, perceive their environment, persist over a prolonged time period, adapt to change, and create and pursue goals" [11, p. 4]. An agent defines its action not for itself but with an environment it interacts with. It recognizes the environment by its sensors, has an agent program to decide what to do with the input data, and performs an action with its actuators. To become a rational agent, the agent must also act to achieve the highest expected outcome according to this performance measure—based on the current and past knowledge of the environment and the possible actions.

When it comes to the general demarcation of agents, according to Russel & Norvig, the agent program can be segmented into four different agent types [11]: A *simple reflex* agent reacts only based on its sensor data whereas a *model-based reflex* agent also considers an internal state of the agent. A *goal-based* agent decides for the best decision to achieve its goals. The fulfilment of a goal is a binary decision which means it can either be fulfilled or not. On the contrast, a *utility-based* agent has no binary goal but a whole utility function which it tries to maximize. An agent can become a *learning agent* by extending its program. Such a *learning agent* then consists of a performance element which selects an action based on the sensor data and a learning element, which gets feedback from the environment, generates own problems, and improves the performance element if possible.

The agent-environment framework consists of three components: an agent, an environment and a goal. Intelligence is the measurement for the "agent's ability to achieve goals in a wide range of environments" [41, p. 12]. The agent gets input by perceptions generated from the environment. One type of perceptions are observations of the environment, while others are reward signals that indicate how well the goals of the agent are achieved. Based on these input signals, the agent decides to perform actions which are sent back as signals to the environment.

# 3. A framework for understanding the role of machine learning in artificial intelligence

In order to understand the interplay of machine learning and AI, we base our concept on the framework of Russel & Norvig [11]. With their differentiation between the two objectives of AI application, *acting* and *thinking*, they lay an important foundation.

## 3.1. Layers of agents

When trying to understand the role of machine learning within AI, we need to take a perspective which has a focus on the implementation of intelligent agents. We require this perspective, as it allows us to map the different tasks and components of machine learning to the capabilities of intelligent agents. If we regard the capabilities of *thinking* and *acting* of an intelligent agent and translate this into the terms of software design, we can reason that the *acting* capabilities can be regarded as a *frontend*, while the *thinking* part can be regarded as a *backend*. Software engineers typically strictly separate form and function to allow for more flexibility and independence as well as to enable parallel development [45]. The frontend is the interface the environment interacts with. It can take many forms. In the case of intelligent agents it can be a very abstract, machine-readable web interface [46], a human-readable application [47] or even a humanoid template with elaborated expression capabilities [48]. For the frontend to interact with the environment, it requires two technical components; sensors and actuators. *Sensors* detect events or changes in the environment and forward the information via the frontend to the backend. For instance, they can read the temperature within an industrial production machine [49] or read visuals of an interaction with a human [50]. Actuators on the other hand are components that are responsible for moving and controlling a mechanism. While sensors just process information, actuators *act*, for instance by automatically buying stocks [51] or changing the facial expressions of a humanoid [52]. One could argue that the Turing test [39] takes place at the interaction of the environment with the frontend, more precisely the combination of sensors and actuators if one wants to test the agent's AI of *acting humanly*. Despite every frontend having sensors and actuators, it is not of importance for our work what the precise frontend looks like; it is only relevant to note that a backend-independent, encapsulated frontend exists.

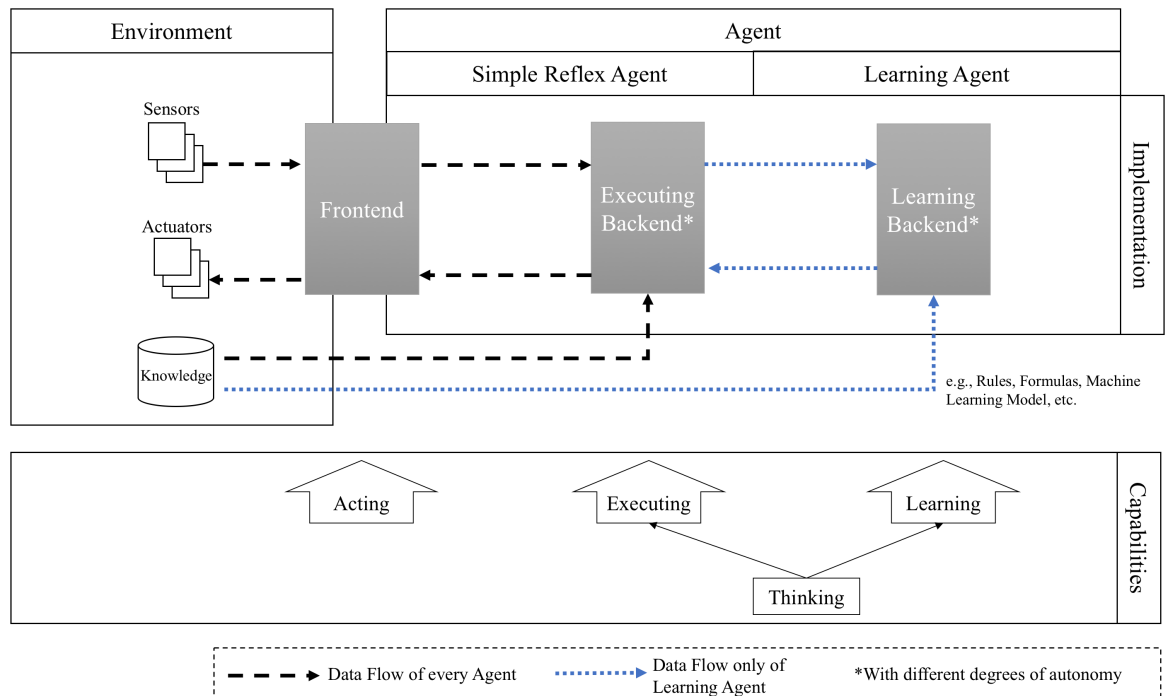

**Figure 2. Conceptual framework**

The backend provides the necessary functionalities, which depict the *thinking* capabilities of an intelligent agent. Therefore, the agent needs to learn and apply learned knowledge.

In consequence, machine learning is relevant in this implementation layer. When regarding the case of supervised machine learning, we need to further differentiate between the process task that is building (=training) adequate machine learning models [21] and the process task that is executing the deployed models [53]. Therefore, to further understand the role of machine learning within intelligent agents, we refine the *thinking* layer of agents into a *learning* sublayer (model building) as well as an *executing* sublayer (model execution)[2]. Hence, we regard the necessary implementation for the learning sublayer as the *learning backend*, while the executing sublayer is denoted by the *executing backend*.

### 3.2. Types of learning

The learning backend dictates first *if* the intelligent agent is able to learn, and, second, *how* the agent is able to learn, e.g., which precise algorithms it uses, what type of data processing is applied, how concept drift [54] is handled, etc. Therefore, we pick up on the terminology from Russel & Norvig [11] by regarding two different types of intelligent agents: *simple-reflex agents* as well as *learning agents*. This differentiation especially holds for a machine learning perspective on AI, as it considers whether the underlying models in the *thinking* layer are once trained and never touched again (simple-reflex)—or continuously updated and adaptive (learning). In recent literature, suitable examples for both can be found. As an example for simple-reflex agents, Oroszi and Ruhland build and deploy an early warning system of pneumonia in hospitals [55]: While building and testing the model for the agent shows convincing results, the adaptive learning of the system after deployment might be critical. Other examples of agents with single-trained models are common in different areas, for instance for anaphora resolutions [56], prediction of pedestrians [57] or object annotation [58]. On the other hand, recent literature also gives examples for learning agents. Mitchell et al. present the concept of "never-ending learning" agents [59] which have a strong focus on continuously building and updating models within agents. An example for such an agent is shown by Liebman et al., who build a self-learning agent for music playlist recommendations [60]. Other cases are for instance the regulation of heat pump thermostats [61], an agent to acquire collective knowledge over different tasks [62] or learning word meanings [63].

---
[2] Russel & Norvig indicate a related relationship by differentiating into *learning elements* and *performance elements* [11].

The choice on this feature in general (simple-reflex vs. learning agent) influences the overall design of the agent as well as the contribution of machine learning. The overview of our resulting framework is depicted in figure 2. In conclusion, in the case of a simple-reflex agent, machine learning takes places as a once-trained model in the execution sublayer. In contrast, it plays a role in the learning sublayer of a learning agent to continuously improve the model in the execution sublayer. This improvement is based on knowledge and feedback, which is derived from the environment via the execution layer.

## 3.3. Continuum between human involvement and machine involvement

When it comes to the executing backend and the learning backend, it is not only of importance if and how underlying machine learning models are updated—but how much automated the necessary processes are. Every machine learning task involves various process steps, including data source selection, data collection, preprocessing, model building, evaluating, deploying, executing and improving (e.g. [21], [53], [64]). While a discussion of the individual steps is beyond the scope of this paper, the autonomy and the automation of these tasks as an implementation within the agent is of particular interest in each necessary task of the machine learning lifecycle [27].

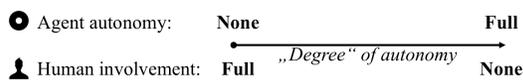

**Figure 3. Degree of agent autonomy and human involvement**

For instance, while the execution of a once-built model can be fairly easily automated, the automated identification of an adequate data source for a new problem or retraining as well as a self-induced model building are more difficult. Therefore, we need to view the human involvement in the necessary machine learning tasks of an intelligent agent, as depicted in figure 3. While it is hard to draw a clear line between all possible forms of human involvement in the machine learning-relevant tasks of an intelligent agent, we see this phenomenon rather as a continuum. The continuum ranges between none or little agent autonomy with full human involvement (e.g. [65]–[67]) on the one extreme as well as the full agent autonomy and no or little human involvement for the delivered task on the other (e.g. [68]–[70]). For example, an intelligent agent with the task to autonomously drive a car considering the traffic signs already proves a high degree of agent autonomy. However, if the agent is confronted with a new traffic sign, the learning of this new circumstance might still need human involvement as the agent might not be able to "completely learn by itself" [71]. Therefore, the necessary involvement of humans, especially in the *thinking* layer (= executing backend and learning backend), is of major interest when describing AI and the underlying machine learning models. The degree of autonomy for each step of machine learning can be investigated and may help to characterize the autonomy of an agent in terms of the related machine learning tasks.

## 4. Research priorities for machine-learning-enabled artificial intelligence

The presented framework of machine learning and its role within intelligent agents is still on a conceptual level. However, given the misunderstandings and ambiguity of the two terms [6–9], we see potential for further research with the aim both to clarify the terminology and to map uncharted territory for machine-learning enabled artificial intelligence.

First, empiric validation as well as continuous, iterative development of the framework is necessary. We need to identify various cases of intelligent agents across different disciplines and to evaluate how well the framework fits. It would be interesting to see how practical and academic machine-learning-enabled artificial intelligence projects map to the framework, and, furthermore even quantify which share of such projects works with learning agents and which with non-learning agents. Additionally, such cases would help us to gain a better understanding of the necessary human involvement in state-of-the art intelligent agents—and, therefore, determine the "degree" of autonomy when regarding all aspects (acting, executing, learning) of such agents.

Second, one aspect of interest would be to reduce the necessary involvement of humans. As stated before, we see this spectrum as a continuum between human involvement and agent autonomy. Two possibilities come immediately to mind. The methods of *transfer machine learning* deal with possibilities on how to transfer knowledge (i.e., models) from one source environment to a target environment [72]. This could indeed help to minimize human involvement, as further research in this field could show possibilities and application-oriented techniques to utilize transfer machine learning for automated adaption of novel or modified tasks [73].

Additionally, regarding already deployed models as part of the backend-layer, it is of interest not only how the models are built initially, but how to deal with changes in the environment. The so-called subfield of *concept drift* holds many possibilities on how to detect changes and adapt models—however, fields of successful application remain rare [54], [74].

## 5. Conclusion

In this paper, we clarify the role of machine learning within artificial intelligence—in particular intelligent agents. We present a framework, which highlights the two cases of simple-reflex and learning agents as well as the role machine learning can play in each of them. In a nutshell, machine learning models can be implemented as once-trained models within an intelligent agent—without the possibility to learn additional insights from the environment (simple reflex agent). Implementation-wise, we call this sublayer of executing knowledge the *executing backend*. In this case, the agent is able to utilize (previously built) machine learning models—but not build and update its own ones. If the agent, however, is able to learn from its environment and is, therefore, able to update the machine learning models within the execution sublayer, it is a learning agent. Learning agents have an additional sublayer, the *learning backend*, which allows them to utilize machine learning in terms of model building/training.

When it comes to the implementation of these two sublayers, it is of importance to capture the *degree of autonomy* that the machine learning within the agent requires. This aspect focusses on the human involvement in the necessary machine learning tasks, e.g. the data collection or the choice of an algorithm.

The research at hand is still in a conceptual state and has certain limitations. First, while the proposed framework allows to deepen the understanding of machine learning within AI, empirical studies are still required to see how well existing machine-learning-enabled AI applications fit into this scheme. Expert interviews with AI designers could validate the model and complete and evaluate the level of detail. Furthermore, we need to find ways to *quantify* the human involvement in machine-learning related tasks within AI to gain better understanding of the degree of autonomy of state-of-the-art agents.

Although at an early stage, our framework should allow scientists and practitioners to be more precise when referring to machine learning and AI. It highlights the importance of not using the terms interchangeably but making clear which role machine learning plays within a specific agent implementation.